\documentclass[sigconf]{acmart}
\usepackage[T1]{fontenc}
\usepackage[latin1]{inputenc}
\usepackage{subfig}
\usepackage{float}
\usepackage{graphicx}
\usepackage{babel}
\usepackage{url}
\usepackage[ruled,vlined]{algorithm2e}
\usepackage{caption}

\AtBeginDocument{%
  \providecommand\BibTeX{{%
    \normalfont B\kern-0.5em{\scshape i\kern-0.25em b}\kern-0.8em\TeX}}}

\setcopyright{acmcopyright}
\copyrightyear{2021}
\acmYear{2021}
\acmDOI{10.1145/1122445.1122456}

\acmConference[Oakridge '21]{Oakridge '21: International Conference on Neuromorphic Systems}{July 27--29, 2021}{Oakridge, IL}
\acmBooktitle{Oakridge, IL '21: International Conference on Neuromorphic Systems,
  July 27--29, 2021, Oakridge, IL}
\acmPrice{15.00}
\acmISBN{978-1-4503-XXXX-X/18/06}

\begin{document}

\title{Tightening the Biological Constraints\\ on Gradient-Based Predictive Coding}

\author{Nick Alonso}
\authornote{Both authors contributed equally to this research.}
\email{nalonso2@uci.edu}
\orcid{1234-5678-9012}
\author{Emre Neftci}
\authornotemark[1]
\email{eneftci@uci.edu}
\affiliation{%
  \institution{University of California, Irvine}
  \streetaddress{P.O. Box 1212}
  \city{Irvine}
  \state{CA}
  \country{USA}
  \postcode{92617}
}

\renewcommand{\shortauthors}{Alonso and Neftci}

\begin{abstract}
Predictive coding (PC) is a general theory of cortical function. The local, gradient-based learning rules found in one kind of PC model have recently been shown to closely approximate backpropagation \cite{whittington2017approximation, millidge2020predictive, song2020can}. This finding suggests that this gradient-based PC model may be useful for understanding how the brain solves the credit assignment problem. The model may also be useful for developing local learning algorithms that are compatible with neuromorphic hardware. In this paper, we modify this PC model so that it better fits biological constraints, including the constraints that neurons can only have positive firing rates and the constraint that synapses only flow in one direction. We also compute the gradient-based weight and activity updates given the modified activity values. We show that, under certain conditions, these modified PC networks perform as well or nearly as well on MNIST data as the unmodified PC model and networks trained with backpropagation.

\end{abstract}




\maketitle

\section{Introduction}

Predictive coding (PC) is a general theory of the function of top-down and bottom-up processing in the neocortex \cite{Rao_Ballard99_predcodi, spratling2008predictive, friston2009predictive}. According to the theory, a central function of top-down projections in the cortex, which are connections that lead from higher to lower level areas of the cortical hierarchy, is to predict neural activity. Differences between the predictions and the actual activity is computed and encoded in error neurons, which feed the errors back up the cortical hierarchy from lower to higher level areas. Learning dynamics and neural activity updates both seek to minimize the prediction errors. While empirical testing of PC theory is needed to validate it, the theory has seen significant empirical support and is consistent with much of what we know about neuroanatomy \cite{huang2011predictive, kok2015predictive, walsh2020evaluating}, making it a potential basis for a general theory of how the neocortex learns and performs inference.

Recent computational work on PC theory has shown that, under certain conditions, a seminal model of PC, originally developed by Rao and Ballard \cite{rao1999predictive}, learns in a way that is very similar to, or in some cases exactly similar to, the gradient back-propagation of errors algorithm \cite{whittington2017approximation, millidge2020predictive, song2020can}. This finding is significant from the point of view of neuroscience because PC networks learn using local, Hebbian-like learning rules, and therefore PC theory could provide a basis for an explanation of how the cortex solves the credit assignment problem. The theory may also advance neuromorphic computing by providing a path toward local learning algorithms that are compatible with neuromorphic hardware \cite{davies2018loihi, qiao2015reconfigurable, friedmann2016demonstrating}.

\begin{figure*}[t]
  \includegraphics[width=15cm]{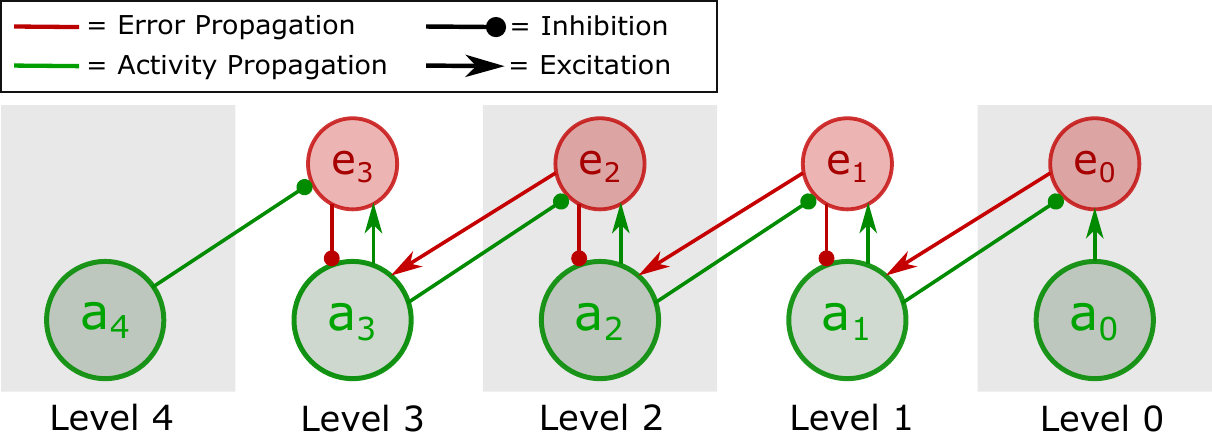}
  \caption{Standard Predictive Coding Model}
  \Description{}
  \label{fig:teaser}
\end{figure*}

In this work, we modify Rao and Ballard's \cite{rao1999predictive} PC model with the goal of making the model more tightly constrained by biology. Doing so will yield both a model that better fits with neurobiology and will be a step toward implementing gradient-based PC algorithms in spiking neural networks. We make and test several modifications to the model. First, the weights that propagate errors in the original PC model are transposes of the forward weights. To avoid this biologically implausible transport of weights, we replace the weight transpose with a separate weight matrix. We test both using the weight matrix untrained (random feedback) and trained using a rule from \cite{kolen1994backpropagation}. Second, the neurons in Rao and Ballard model encode continuous valued firing rates and the values of activity neurons (green nodes in figure \ref{fig:teaser}) are allowed to be negative. Firing rates can only be positive in artificial spiking neurons and biological neurons, so we prevent values from going negative using a ReLU activation function and develop a method for preventing the loss of gradient information due to the ReLU function. Third, it is necessary to report negative errors, but neural activities cannot be negative. We discuss several ways errors can be encoded in spiking neurons with only (positive) firing rate values. We also compute the gradients for these error encoding schemes for firing rate neurons and use them to update the model. 

We test these modifications on a supervised learning task using the MNIST dataset. We find that certain versions of the modified PC model works as well as the original PC model and backpropagation. This suggests the PC model may be a favorable basis on which to develop more neuromorphic and biologically-constrained spiking neural network models that use gradient-based learning.

\section{Gradient-Based Predictive Coding}\label{sec:GradPC}

\begin{table}[h]
 \begin{tabular}{||c c||} 
 \hline
 Term & Value \\ [.7ex] 
 \hline\hline
 $a_l$ & activity at level $l$ \\ 
 \hline
 $e_l$ & prediction error at level $l$ \\
 \hline
 $p_l$ & prediction at level $l$ \\
 \hline
 $W_l$ & synaptic weights from $l$ to $l+1$  \\ [1ex] 
 \hline
\end{tabular}
\end{table}

In this section, we describe the standard gradient-based PC model based on Rao and Ballard's model \cite{rao1999predictive}.  We choose to focus on this model because, under certain conditions, the model was demonstrated to closely approximate gradient backpropagation \cite{whittington2017approximation, millidge2020predictive, song2020can}. Song et al. \cite{song2020can}, for example, showed that the first non-zero weight update that occurs to each weight matrix during supervised training is exactly identical to those made by backpropagation. Whittington et al. \cite{whittington2017approximation} showed that after the network converges, the equations that define how errors are propagated through the network match those of the backpropagation algorithm. This relation is important because it provides a formal link between PC and the minimization of a global objective function through gradient ascent, which is a method that works well in artificial neural networks (ANN) at scale. Other predictive coding that lack this connection have no guarantees on their ability to minimize global objectives.

There are two kinds of neurons in the  PC model: activity neurons and prediction error neurons. Activity neurons represent features of the input, while error neurons encode differences between predictions of activity neurons and the actual values of activity neurons. Predictions of activities at level $l$ are passed from level $l-1$ according to the following equation:
\begin{equation}\label{eq:spc_pred_neuron}
p_l = W_{l-1}a_{l-1}.
\end{equation}
A local prediction error is then computed and encoded in error neurons. The error is the difference between the activity and the prediction, which is first passed through an element-wise non-linearity $f$.
\begin{equation}\label{eq:spc_err_neuron}
e_l = a_l - f(p_l).
\end{equation}
The weight matrices $W_l$ are trained using this local error according to: 
\begin{equation}\label{eq:spc_w_update}
\begin{split}
W_l \leftarrow  \alpha \Delta W_l + W_l,\text{ where }
\Delta W_l &= e_{l+1} f'(p_{l+1}) a_{l}^T.
\end{split}
\end{equation}

$\Delta W_l$ is multiplied by a small learning rate $\alpha$ before being added to the weights, where $0 < \alpha < 1$. This update is equivalent to taking the gradient of the squared prediction error $\frac12 e_l^2$ with respect to the weights that produced the prediction \cite{rao1999predictive}.

In addition to updating their weights,  PC networks also optimize their activities. Activities optimize the same cost, i.e. the squared prediction errors at each level. At each time-step, they update their activities according to:
\begin{equation}\label{eq:spc_a_update}
\Delta a_l = W_{l}^T e_{l+1} f'(p_{l+1}) - e_{l}.
\end{equation}
Here, $f'$ is the gradient of the non-linearity. Updating activities at level $l$ is equivalent to performing gradient ascent over the activities using the gradient of the squared prediction error at level $l$ and the next level $l+1$. Because one is performing gradient ascent, the computed gradient must be multiplied by a small inference rate $\beta$, where $0 < \beta < 1$, before adding to the activities. One typically wants the activities to update quicker than the weights, so $\beta$ should be larger than $\alpha$. This is because one wants activities to converge quickly to an optimal representation of the input, while learning (i.e. weight change) should occur slowly to ensure generalization ability.  

\section{Predictive Coding Constrained}

We now describe our modified PC network. Each modification to the network is made to make the model fit better with neurobiology. However, we wish to make these changes without losing the ability of the network to propagate gradients and perform similarly to backpropagation. 

\subsection{Predictive Coding without Weight Transport}
Equation \ref{eq:spc_a_update} shows that errors propagated back through the network require using the transpose of the forward weights. 
Using the transpose of forward weights to propagate errors is generally considered biologically implausible as it implies a bidirectional synapse. 
Transposing a weight matrix is also problematic from a neuromorphic hardware implementation perspective because it requires the duplication and synchronization of the weights on the pre-synaptic and the post-synaptic side. 

The weight transport problem has been tackled in the context of gradient-backpropagation for conventional deep networks using approximations such as random feedback weights \cite{lillicrap2016random} and local loss functions \cite{mostafa2018deep,akrout2019deep, kunin2020two}. These approximations underperform slightly compared to exact gradient back-propagation, but do not require a symmetric transpose of the network weights.

We propose two modifications to PC networks that avoid this problem, which build off of previous work done on the weight transport problem for backpropagation. 

In the first network model, which we call Rand-PC, we replace the $W^T$ in equation \ref{eq:spc_a_update} with a random matrix $B$ of the same size. Replacing weight transposes with random feedback matrices has been shown to work reasonably well for approximating backpropagation in conventional deep networks \cite{lillicrap2016random, xiao2018biologically}. In the results section, we show it works nearly as well as backpropagation in our predictive coding networks.

In the second network model, which we call the \textit{Kollen-Pollack PC} (KP-PC), we use a method developed by Kolen and Pollack \cite{kolen1994backpropagation}, and extended by Akrout et al. \cite{akrout2019deep}, to train backward matrices so that they converge to the transposes of the forward matrices. 
Like Rand-PC, with KP-PC we replace the weight transpose in equation \ref{eq:spc_a_update} with a separate weight matrix $B_l$. This weight matrix is then trained with the transpose of the update to the forward $W_l$.
Kollen and Pollack build on the simple idea that, if one makes the same weight update to two matrices, $W$ and $B$, the two matrices will grow more similar. However, due to numerical precision errors, the two matrices in practice eventually diverge. To solve this problem in conventional ANN, Kolen and Pollack add a small decay term to both updates. They show that $W$ and $B$ will eventually converge to (nearly) the same matrix.
\begin{equation}
\begin{split}
\Delta W_l &=  A_l - \gamma W_l,\,\Delta B_l =  A_l^T - \gamma B_l,\\
\end{split}
\end{equation}
where $0<\gamma<1$ is a small decay rate and $A_l$ is the weight adjustment. Acrout et al. \cite{akrout2019deep} point out that these weight updates are local for the forward and backward matrices, if each weight matrix is connected to the same two populations of neurons. Interestingly, as can be seen from the equations in section \ref{sec:GradPC} and figure \ref{fig:teaser}, PC networks require just this structure, where forward weights take input from activities at level $l$ and output to error neurons at level $l+1$, while backward weights to the opposite. 

\subsection{Biologically Constrained Activity Neurons}

Equation \ref{eq:spc_a_update} shows that PC networks allow for negative neural activity values because updates are linear. 
In biological neurons and artificial spiking neurons, an activation $a$ typically represents the firing rate of a neuron, which cannot be negative. Thus, in a neural modelling and neuromorphic computing setting it is desirable to have neurons with non-negative activities. This raises the question of how well PC models work under the constraint that activities are non-negative. To test this, we enforce positive activation values by passing the activities through a rectified linear unit (ReLU) function after each activity update to ensure there are no negative values.

Additionally, to help prevent a loss of gradient information due to the ReLU function, we add a bias term to the model. In equation \ref{eq:spc_err_neuron}, for example, the error $e_l$ is now equal to $a_l - f(p_l) + b$. For the subtraction and division threshold schemes, one simply replaces $f(p_l)$ with $f(p_l) + b$ in the equations where $f(p_l)$ is present. No other changes are needed. This simple addition to the model will help prevent activities from going negative and will therefore help prevent gradient information from being lost when negative activities are zeroed out. The bias will do this by forcing predictions to be greater than zero. This will make sure the network activities are initialized to non-negative values. It will also help activities maintain non-negative during activity updates, through the top-down error (second term in equation \ref{eq:spc_err_neuron}). The top-down error pulls the activities toward the predictions at the same level. If predictions are always non-zero, they will always pull activities toward non-zero values. If activities a non-negative, the gradients they accumulate will not be lost when passed through a ReLU function.

\subsection{Biologically Constrained Error Neurons}

Like the activity neurons, error neurons, according to equation 2, can take on negative values when activity values are over-predicted. For the same reasons listed in the previous section, it is desirable to have error neurons with only positive firing rate values. This raises the question of how prediction errors could be encoded in spiking patterns with non-negative firing rates.

Rao and Ballard and several other scientists (e.g.\cite{keller2018predictive}) hypothesize there could be two kinds of prediction error neurons in the brain. One kind spikes in response to over-predicted values (i.e. negative errors according to equation \ref{eq:spc_err_neuron}). The other kind spikes in response to under-predicted values (i.e. positive errors according to equation \ref{eq:spc_err_neuron}). This encoding scheme implies that generally, when prediction errors are large the error neuron activities will be large, and while errors are small error neuron activity is small. These are the sort of error neurons that are assumed to exist in the standard PC model. Implementing this encoding scheme could, for example, use one set of error neurons that encode only the positive values of $a_l - f(p_l)$ (underpredictions) and another set of error neurons encoding only the positive values of $f(p_l) - a_l$ (overpredictions). We will call this sort of error encoding scheme \textit{subtractive separated encoding} since the subtractive error is encoded in two separate kinds of error neurons.

An alternative way to encode errors in spike trains involves setting error neurons to have a baseline activity rate, which is dampened when activities are over-predicted and excited when they are under-predicted. When subtraction is used to dampen and excite error neurons, we call this encoding scheme the \textit{subtractive threshold encoding scheme}. There may be several ways to implement this in spiking neurons. For example, one could use a constant input of current that causes the error neurons to fire at a certain rate even when errors are absent.

Here we show how a subtractive threshold scheme can be developed in firing rate neurons so weight and activity updates are equivalent to the original Rao and Ballard equations. Consider the following error neuron encoding scheme
\begin{equation}
e_l^* = \frac{2}{e_{\max}}((a_l - f(p_l) - e_{\min} ) = \frac{2 (e_l - e_{\min} )}{e_{\max} }.
\end{equation}
Here, $e_{\min}$ is the minimum possible value of $e_l$, and $e_{\max}$ is the maximum possible value of $e_l - e_{\min}$. There will be a minimum value if activities are clamped to have a minimum value and $f$ is a squashing non-linearity (e.g. sigmoid). If there is no maximum one can replace $e_{\max}$ with an approximate $e_{\max}$ value to obtain a similar result. For example, we apply the sigmoid function to $p_l$, and the ReLU function to the activities. If activities never go above 1.1, then $e_{\min} = -1$ and $e_{\max} = 2.1$. This equation forces all error values to be between 0 and 2, with a baseline firing rate of 1. One can replace the 2 with another value which determines the maximum firing rate. 

In spiking neurons, the $e_{\min}$ can be thought of as a constant excitatory current source that, absent any other inputs, causes the neurons to spontaneously fire at a constant rate. 
The $\frac{2}{e_{\max}}$ term can also be seen as a constant input current that performs some form of normalization, which is well known to be a pervasive computation throughout the brain \cite{carandini2012normalization}.

Weight and activity updates are computed by replacing the $e_l$ term in equation \ref{eq:spc_err_neuron} and \ref{eq:spc_a_update} with $(\frac{e_{\max}}{2} e_l^* + e_{\min} )$, since $e_l = (\frac{e_{\max}}{2} e_l^* + e_{\min} )$. With these replacements, updates remain local (since they only depend on the local errors, local activities, and constants) and are equivalent to the original equations.

We note that these two error encoding schemes with biological or spiking neurons can be obtained as special cases of population decoding \cite{dayan2001theoretical}, which can be conveniently realized with e.g. the neural engineering framework \cite{eliasmith2003neural}. 
Our focus on these special cases is due to their efficiency, as they require one or two neurons per encoded error value compared to population of neurons per error value in the general case. 

A third way to encode mismatches between predictions and activities is an encoding scheme that involves dividing the activities by the predictions (or vice versa). Spratling \cite{spratling2008predictive}, for example,  developed a firing rate model of predictive coding which uses a division error term that divides activities (element-wise) by the predictions. His model was able to replicate some neurophysiological data, including fine-grained calcium imaging data that seemed to show the existence of neurons in the mouse primary visual cortex that were sensitive to mismatches between actual and predicted visual flow \cite{spratling2019fitting}. We call encoding schemes of this form \textit{division mismatch encoding}. 

Spratling's particular model is not formulated to minimize a global loss using its gradients w.r.t. weights and activities. We are interested in working within the framework of gradient ascent as it has proven effective in large neural networks. We thus present an alternative firing rate model with division mismatch encoding that maintains gradient updates on activities and errors. We compute values of the mismatch neurons as follows:
\begin{equation}
e_l^{**} = \sqrt{\frac{a_l + \epsilon}{f(p_l) + \epsilon}}.
\end{equation}
The $\epsilon$ is a small constant that prevents division by zero and helps prevent exploding gradients. The term $\frac{a_l+ \epsilon}{f(p_l) + \epsilon}$ ranges between 0 and $\infty$ (assuming activities and predictions are positive). Over-predictions range between $0$ and $1$, while under-predictions range between $1$ and $\infty$. We find that, although not necessary, adding the square root function improves learning slightly and makes activity updates more stable. When activities equal the predictions, $e_l^{**}$ will equal one, so we develop a cost function that measures the difference between $e_l^{**}$ and $1$ (see appendix). Weights and activities are then updated in proportion to the gradient of this new cost function (see appendix).

\begin{algorithm} 
\SetAlgoLined
\DontPrintSemicolon
\KwData{($X_B, Y_B$) = dataset X and targets Y formed into minibatches B \textbf{}} 
 
 \Begin{
    \tcp{For each minibatch perform n activity updates then a weight update}
    \For{$(x_b,y_b) \in (X_B, Y_B)$}{
        \tcp{Initialize}
        $a_0 \leftarrow x_b$\;
        \For{$l=1$ \KwTo $L$}{
            $p_l \leftarrow W_{l-1}a_{l-1}$\;
            $a_l \leftarrow f(p_l)$
        }
        $a_L \leftarrow y_b$\;
        \;
        \tcp{Perform $n$ Activity Updates}
        \For{$i=0$ \KwTo $n$}{
            \For{$l=1$ \KwTo $L$}{
                \tcp{Compute Errors}
                equation 2, 6, or 7\;
                \tcp{Activity Update}
                equation 4\;
                \tcp{Prediction Update}
                equation 2\;
            }
        }
        \;
        \tcp{Train Weight Matrices}
        \For{$l=0$ \KwTo $L-1$}{
            \tcp{Compute Errors then update weights}
            equation 2, 6, or 7\;
            equation 3\;
        }
    }
  }

 \caption{Supervised Predictive Coding Training \label{alg:1}}
\end{algorithm}

\section{Results}

Predictive coding networks can be trained for both supervised and self-supervised learning tasks. In self-supervised learning tasks, the activities of the bottom level activities (level 0 in figure \ref{fig:teaser}) are set to the data (e.g. image), while the activities of the top level (level 4 in figure \ref{fig:teaser}) are set to some constant (e.g. 1). Alternatively, the top level can be removed entirely. The network weights can then be trained online after each update to the activities, or a single weight update can occur after the network activities converge \cite{bogacz2017tutorial}. 

In what follows, we show the results of training our models using supervised methods, which were developed in \cite{whittington2017approximation}. In supervised learning, the bottom level activities of the PC network are clamped to the target, while the top-level activities are clamped to the input data. Activities are initialized by propagating predictions down level by level to level 0. Activities of hidden layers are set to the predictions at each level: $a_l^{initial} = f(p_l^{initial})$. After initialization, activities are optimized. Weight updates can occur online or after the activities reach convergence (See Algorithm \ref{alg:1}). We found activity convergence is slow for MNIST, so instead of waiting for convergence we update weights after $20$ activity updates. For fashion-MNIST, the activities converge more quickly, so we update weights after $7$ activity updates. When tested, the the activities at level $4$ are clamped to the test image, then predictions propagate down to the output layer (level $0$), where the output is compared to the target.

Inference rates are set to .1 for MNIST and .025 for Fashion-MNIST. Learning rates are set to .001. Adam optimizers are used to train all weight matrices. All models use a network with fully connected layers of size of 784-300-300-10. We train the networks to classify images in the MNIST and Fashion-MNIST data set. Each dataset consists of gray-scale images of size 28x28. There are 60,000 training images and 10,000 test images in each dataset. We show the classification accuracies on the test sets below. \footnote{Code for models can be found here: \url{https://github.com/nalonso2/Tightening-the-Biological-Constraints-on-Gradient-Based-Predictive-Coding}}

\subsection{PC with Separate Feedback Weights}

\begin{figure}[h]
  \includegraphics[width=\linewidth]{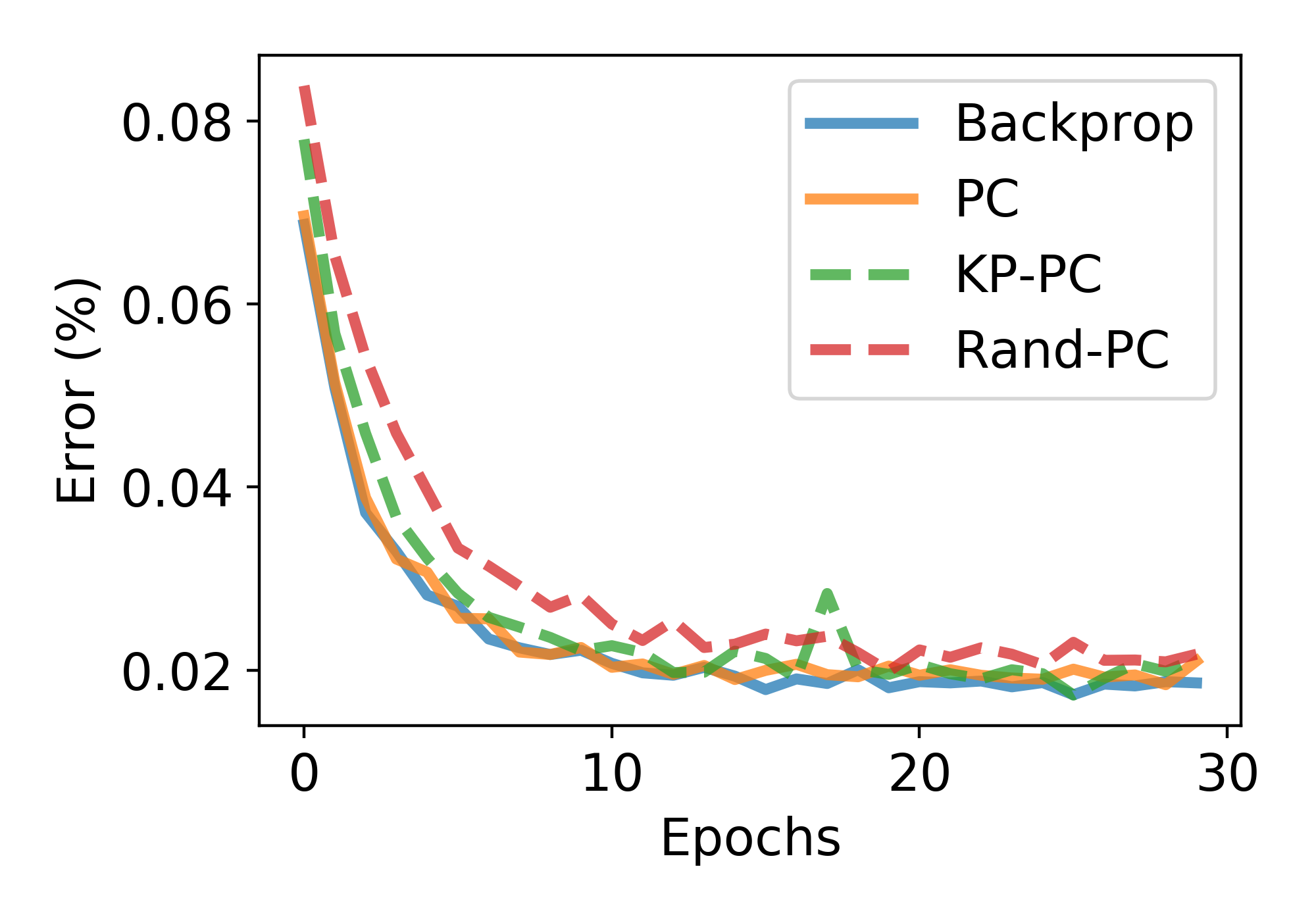}
  \caption{Test errors of PC models with various feedback weights on MNIST with sigmoid activations.}
  \Description{}
  \label{fig:accweights}
\end{figure}

We begin by testing how well PC networks train with the separate feedback matrices discussed in section 2.3. Here all networks use sigmoid activation functions. Rand-PC uses a fixed, random feedback matrix, while the KP-PC uses a weight matrix trained using the rules discussed in section 2.3. The test accuracy, averaged over three runs, is shown after each epoch of training starting after the first training epoch. 

\begin{table*}
  
  \begin{tabular}{c c c c c c c}
    \toprule
    Data & Backprop & PC & KP-PC & Rand-PC & PC w/ Div & Rand-PC w/ Div\\
    \midrule
    MNIST & $.019$ $(1x10^{-3})$ & $.020$ $(1x10^{-3})$ & $.019$ $(2x10^{-3})$ & $.021$ $(1x10^{-3})$ & $.021$ $(8x10^{-4})$ & $.022$ $(1x10^{-3})$\\
    \midrule
    Fashion-MNIST & $.107$ $(2x10^{-3})$ & $.109$ $(3x10^{-3})$ & $.116$ $(5x10^{-3})$ & $.112$ $(3x10^{-3})$ & $.116$ $(4x10^{-3})$ & $.116$ $(2x10^{-3})$\\
    \bottomrule

  \end{tabular}
  \captionsetup{width=.86\linewidth}
  \caption{Mean (and standard deviation) of validation errors (\%) for  PC models with sigmoid activitations. Three different seeds of each model were trained. Values shown are the means and standard deviations of the validation errors of the last three epochs across training runs.}
  \label{tab:acc}
  
\end{table*}

We find the standard PC network, which uses the weight transpose for feedback, performs as well as backpropagation, which replicates the findings of \cite{whittington2017approximation}. The KP-PC network also performs as well as backpropagation, while the Rand-PC network only does slightly worse (see table \ref{tab:acc}). All models achieve a mean accuracy within a standard deviation of $98\%$. Similar results are found with fashion-MNIST, where these models' mean test errors were within two standard deviations of backpropagation.

\subsection{PC with Constrained Activity Rates}

Next, we test how well PC networks work under the constraint that activity neurons can only take positive firing rate values. As mentioned above, because firing rates are updated linearly, activity values can become negative even when initialized to be positive. It is possible then, that preventing activities from being negative (by passing them through a ReLU function after each update) may erase gradient information useful for credit assignment, and this may consequently hurt performance. Here we test how severe this potential loss of gradient information is when using different activation functions. All models have sigmoid activations at the output layer, but some models use sigmoid functions at hidden layers while others use tanh functions at hidden layers. 

\begin{figure}[h]
  \includegraphics[width=\linewidth]{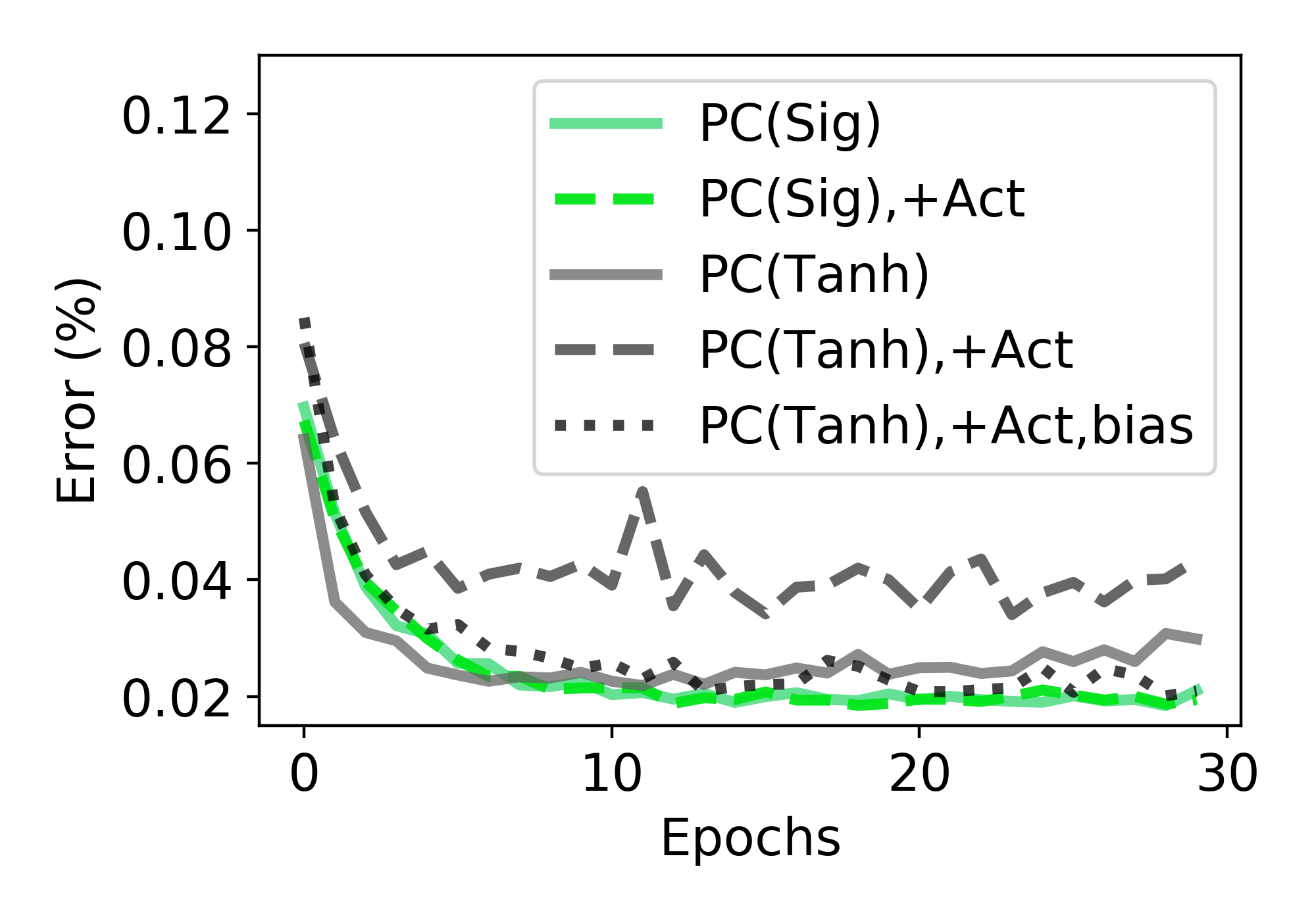}
  \caption{Test errors of PC models on MNIST with activities constrained to be positive (+Act) versus those allowed to go negative.}
  \Description{}
  \label{AccPosActs}
\end{figure}

Figure \ref{AccPosActs} shows that constraining activities to be positive does not negatively affect the performance of the network when sigmoid activation functions are used. However, there is a small but significant drop in performance when Tanh non-linearities are used. When a small bias is added, as specified in section 3.3, the network using Tanh non-linearities sees no drop in performance. 

There are a couple reasons why we see this pattern. When the network is initialized, the activities are set to the predictions which are passed through the non-linearity (see \ref{alg:1}). The same process is used to generate predictions at test time. Because sigmoid maps all values to non-negative numbers, the network, when tested, will have activities with only positive values. Thus, when ReLUs are applied, no activities are affected. This is not true when Tanh is applied without a bias, which will map some numbers to negative values. However, when a bias is added, the Tanh network will not produce negative predictions.

However, this does not explain why learning in the sigmoid network without bias is not affected by constrained activity values, since updates to the activities that occur after the initialization can still push the activities to negative values. If an update pushed activity values to be negative, then the negative values will be set to zero and the gradient information they encode will be lost. This loss of gradient information will presumably negatively affect learning.

We find through experimentation, however, that the top-down error term (the second term in equation \ref{eq:spc_err_neuron}) prevents activities from going too far outside the range of the predictions. When predictions are passed through a sigmoid, for example, the top-down error will pull the activities toward the prediction, which is some positive number between zero and one. This generally prevents activities from going below zero. The same is not true of predictions passed through Tanh without a bias, which will not always pull activities toward positive values. In this case, useful gradient information will be lost when negative activities are clamped to zero. A bias term, thus, is necessary in cases where non-linearities allow for negative predictions, but may not be necessary for non-linearities that map all input to non-negative numbers.

\subsection{Subtraction versus Division Errors}

In section $4.3$, we outlined three error neuron encoding schemes: subtractive separated encoding, subtractive threshold encoding, and division mismatch encoding. We also showed how to compute the gradients for the new encoding schemes. As explained above, the two subtractive error encoding schemes will look quite different in spiking neurons. However, we showed that both of these encoding schemes are mathematically equivalent in the firing rate model, and that the separated subtractive encoding scheme works as well as backpropagation. Here we test how well the division threshold encoding scheme compares to backpropagation.

Positive activity values are required for the division encoding scheme (due to the square-root and logarithm in equations 7 and 8), so we set all models to have positive activity values. We apply sigmoid activation at hidden layers. We still find that adding a small bias is necessary to prevent loss of gradient information, so we add a small bias.

\begin{figure}[h]
  \includegraphics[width=\linewidth]{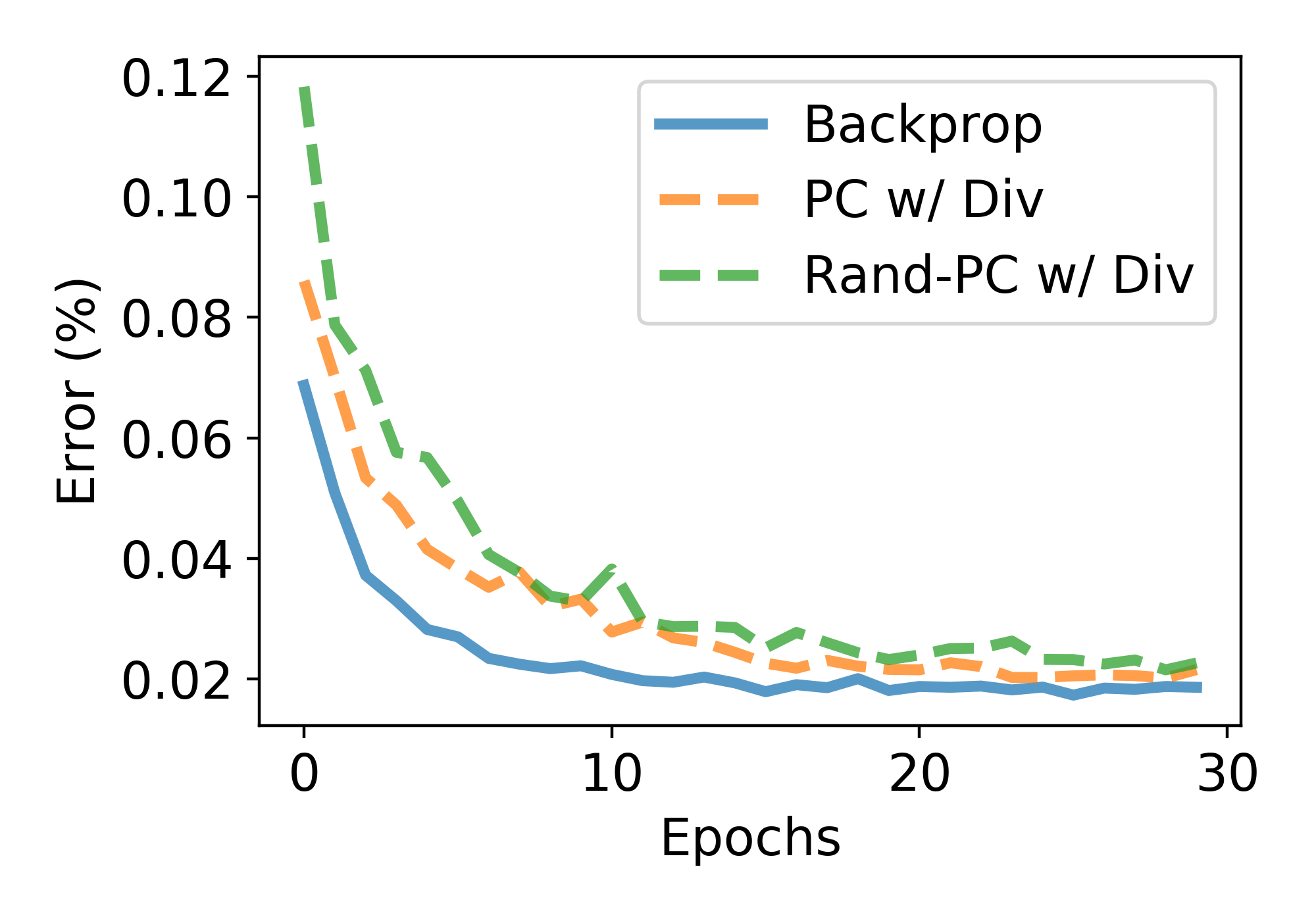}
  \caption{Test errors of PC models on MNIST with division encoding errors.}
  \Description{}
  \label{fig:AccDivErr}
  
\end{figure}

We can see in figure \ref{fig:AccDivErr} that the PC models that use division based errors produce comparable results to backpropagation. The division mismatch model that uses true gradients achieves a mean test error of $.021$ while the division model with random feedback achieves a mean test error of $.22$ (see table \ref{tab:acc}), which is only slightly worse than the performance of backpropagation, which had a mean test error of $.019$. The results are similar for fashion MNIST as well, where both division mismatch models produce mean test errors with less the one percent difference of the mean test error of backpropagation (see \ref{tab:acc}).

\section{Discussion}

In this paper, we showed that a more biologically constrained version of Rao and Ballard's \cite{rao1999predictive} seminal model of predictive coding performed similarly to backpropagation on supervised learning tasks using MNIST data. We found this to be true under constraints where 1) separate feedback weights were used to propagate errors, 2) activity values were prevented from going negative, and 3) error neuron activities were prevented from going negative using either division or subtraction based encoding schemes. We also showed how the gradients for the new encoding schemes could be computed and incorporated into the model.

These results suggest that it is likely possible for more biologically constrained models of gradient-based predictive coding to be built using spiking neural networks. We computed the gradients for the division and subtraction threshold error encoding schemes, which prevent error neurons from having negative activity rates. These equations can potentially be used as a basis for forming new equations for spiking neuron models. We also discussed how separated subtraction error encoding could be implemented in spiking neurons. Although spiking models of PC have been previously demonstrated (e.g. \cite{wacongne2012neuronal}), as far as we can find, spiking neural models of PC that utilize gradient-based inference and learning have yet to be developed.

Of course, in spiking neurons the true gradients cannot be computed because the step functions used as non-linearities are non-differentiable. However, surrogate gradient methods \cite{neftci2019surrogate} used to approximate gradients in spiking neural networks can naturally be incorporated into predictive coding networks. Such surrogate-gradient spiking PC models could help further develop the empirical hypothesis that predictive coding is the general algorithm the brain uses to solve the credit assignment problem. Additionally, such networks could lead to useful local-learning algorithms that are compatible with neuromorphic hardware.

It is still an open question whether the cortex is performing some form of predictive coding. There is good evidence that top-down connections in the cortex do propagate predictions \cite{huang2011predictive, kok2015predictive, walsh2020evaluating}. However, the hypothesis that the cortex encodes prediction errors widely in specialized error neurons (which is a key implication of PC) is not yet widely accepted within neuroscience. There is much data that is consistent with the hypothesis that such error neurons exist in the cortex, but none of it is particularly conclusive (for recent review see \cite{walsh2020evaluating}).

One reason why it has been difficult to locate error neurons is that different error encoding schemes yield different empirical predictions \cite{walsh2020evaluating}. However, some progress is being made. Spratling \cite{spratling2019fitting}, for example, recently found that division-based error encoding schemes better fit certain neurophysiological data than the subtractive encoding of the Rao-Ballard model. Spratling did not compare the physiological data to a subtractive \textit{threshold} encoding scheme, like the one we proposed here (equation 6), so it is unclear whether division error encoding fits with the data better than subtractive schemes generally or only with a particular type of subtractive encoding present in the Rao-Ballard model. Nonetheless, this study shows how we can begin to build evidence in favor of one hypothesis over another.

Our model also illustrates that gradient-based PC is compatible with multiple different kinds of error encoding schemes. So the general hypothesis that some form of gradient-based PC is utilized by the cortex does not depend on there existing a subtractive rather than divisive error encoding. It will, instead, depend on the way the errors are propagated through the cortex and used to affect neural activity and learning. It may be that errors and weights are updated using some other optimization method (e.g. \cite{spratling2009unsupervised}).

More empirical and computational work will be needed to settle these debates. We hope, however, that the work presented here provides new avenues to further develop and test neural models of gradient-based predictive coding and to further develop useful brain-inspired learning algorithms.

\section{Conclusion}

PC is a theory that is of interest to both cognitive science, neuroscience, and engineering. It, for one, is able to explain a wide array of neurophysiological and anatomical data. PC also has potential to provide a path toward understanding how the brain solves the credit assignment problem. We also suggested that the local learning rules used within PC models could lead to useful learning algorithms that are compatible with the constraints neuromorphic hardware. In this paper, we showed that the standard gradient-based PC model can learn and perform inference well under tighter biological constraints. This further supports the position that PC can be developed in a way that is useful for both neuroscience and neuromorphic computing, and marks a path along which PC can be further developed.

\begin{acks}
This work was supported by the National Science Foundation under grant 1652159 and 1823366 (EN).
\end{acks}

\bibliographystyle{ACM-Reference-Format}
\bibliography{biblio_unique_alt,sample-base}


\begin{thebibliography}{28}


\ifx \showCODEN    \undefined \def \showCODEN     #1{\unskip}     \fi
\ifx \showDOI      \undefined \def \showDOI       #1{#1}\fi
\ifx \showISBNx    \undefined \def \showISBNx     #1{\unskip}     \fi
\ifx \showISBNxiii \undefined \def \showISBNxiii  #1{\unskip}     \fi
\ifx \showISSN     \undefined \def \showISSN      #1{\unskip}     \fi
\ifx \showLCCN     \undefined \def \showLCCN      #1{\unskip}     \fi
\ifx \shownote     \undefined \def \shownote      #1{#1}          \fi
\ifx \showarticletitle \undefined \def \showarticletitle #1{#1}   \fi
\ifx \showURL      \undefined \def \showURL       {\relax}        \fi
\providecommand\bibfield[2]{#2}
\providecommand\bibinfo[2]{#2}
\providecommand\natexlab[1]{#1}
\providecommand\showeprint[2][]{arXiv:#2}

\bibitem[\protect\citeauthoryear{Akrout, Wilson, Humphreys, Lillicrap, and
  Tweed}{Akrout et~al\mbox{.}}{2019}]%
        {akrout2019deep}
\bibfield{author}{\bibinfo{person}{Mohamed Akrout}, \bibinfo{person}{Collin
  Wilson}, \bibinfo{person}{Peter~C Humphreys}, \bibinfo{person}{Timothy
  Lillicrap}, {and} \bibinfo{person}{Douglas Tweed}.}
  \bibinfo{year}{2019}\natexlab{}.
\newblock \showarticletitle{Deep learning without weight transport}.
\newblock \bibinfo{journal}{\emph{arXiv preprint arXiv:1904.05391}}
  (\bibinfo{year}{2019}).
\newblock


\bibitem[\protect\citeauthoryear{Bogacz}{Bogacz}{2017}]%
        {bogacz2017tutorial}
\bibfield{author}{\bibinfo{person}{Rafal Bogacz}.}
  \bibinfo{year}{2017}\natexlab{}.
\newblock \showarticletitle{A tutorial on the free-energy framework for
  modelling perception and learning}.
\newblock \bibinfo{journal}{\emph{Journal of mathematical psychology}}
  \bibinfo{volume}{76} (\bibinfo{year}{2017}), \bibinfo{pages}{198--211}.
\newblock


\bibitem[\protect\citeauthoryear{Carandini and Heeger}{Carandini and
  Heeger}{2012}]%
        {carandini2012normalization}
\bibfield{author}{\bibinfo{person}{Matteo Carandini} {and}
  \bibinfo{person}{David~J Heeger}.} \bibinfo{year}{2012}\natexlab{}.
\newblock \showarticletitle{Normalization as a canonical neural computation}.
\newblock \bibinfo{journal}{\emph{Nature Reviews Neuroscience}}
  \bibinfo{volume}{13}, \bibinfo{number}{1} (\bibinfo{year}{2012}),
  \bibinfo{pages}{51--62}.
\newblock


\bibitem[\protect\citeauthoryear{Davies, Srinivasa, Lin, Chinya, Cao, Choday,
  Dimou, Joshi, Imam, Jain, et~al\mbox{.}}{Davies et~al\mbox{.}}{2018}]%
        {davies2018loihi}
\bibfield{author}{\bibinfo{person}{Mike Davies}, \bibinfo{person}{Narayan
  Srinivasa}, \bibinfo{person}{Tsung-Han Lin}, \bibinfo{person}{Gautham
  Chinya}, \bibinfo{person}{Yongqiang Cao}, \bibinfo{person}{Sri~Harsha
  Choday}, \bibinfo{person}{Georgios Dimou}, \bibinfo{person}{Prasad Joshi},
  \bibinfo{person}{Nabil Imam}, \bibinfo{person}{Shweta Jain}, {et~al\mbox{.}}}
  \bibinfo{year}{2018}\natexlab{}.
\newblock \showarticletitle{Loihi: A neuromorphic manycore processor with
  on-chip learning}.
\newblock \bibinfo{journal}{\emph{Ieee Micro}} \bibinfo{volume}{38},
  \bibinfo{number}{1} (\bibinfo{year}{2018}), \bibinfo{pages}{82--99}.
\newblock


\bibitem[\protect\citeauthoryear{Dayan and Abbott}{Dayan and Abbott}{2001}]%
        {dayan2001theoretical}
\bibfield{author}{\bibinfo{person}{Peter Dayan} {and}
  \bibinfo{person}{Laurence~F Abbott}.} \bibinfo{year}{2001}\natexlab{}.
\newblock \bibinfo{booktitle}{\emph{Theoretical neuroscience: computational and
  mathematical modeling of neural systems}}.
\newblock \bibinfo{publisher}{Computational Neuroscience Series}.
\newblock


\bibitem[\protect\citeauthoryear{Eliasmith and Anderson}{Eliasmith and
  Anderson}{2003}]%
        {eliasmith2003neural}
\bibfield{author}{\bibinfo{person}{Chris Eliasmith} {and}
  \bibinfo{person}{Charles~H Anderson}.} \bibinfo{year}{2003}\natexlab{}.
\newblock \bibinfo{booktitle}{\emph{Neural engineering: Computation,
  representation, and dynamics in neurobiological systems}}.
\newblock \bibinfo{publisher}{MIT press}.
\newblock


\bibitem[\protect\citeauthoryear{Friedmann, Schemmel, Gr{\"u}bl, Hartel, Hock,
  and Meier}{Friedmann et~al\mbox{.}}{2016}]%
        {friedmann2016demonstrating}
\bibfield{author}{\bibinfo{person}{Simon Friedmann}, \bibinfo{person}{Johannes
  Schemmel}, \bibinfo{person}{Andreas Gr{\"u}bl}, \bibinfo{person}{Andreas
  Hartel}, \bibinfo{person}{Matthias Hock}, {and} \bibinfo{person}{Karlheinz
  Meier}.} \bibinfo{year}{2016}\natexlab{}.
\newblock \showarticletitle{Demonstrating hybrid learning in a flexible
  neuromorphic hardware system}.
\newblock \bibinfo{journal}{\emph{IEEE transactions on biomedical circuits and
  systems}} \bibinfo{volume}{11}, \bibinfo{number}{1} (\bibinfo{year}{2016}),
  \bibinfo{pages}{128--142}.
\newblock


\bibitem[\protect\citeauthoryear{Friston and Kiebel}{Friston and
  Kiebel}{2009}]%
        {friston2009predictive}
\bibfield{author}{\bibinfo{person}{Karl Friston} {and} \bibinfo{person}{Stefan
  Kiebel}.} \bibinfo{year}{2009}\natexlab{}.
\newblock \showarticletitle{Predictive coding under the free-energy principle}.
\newblock \bibinfo{journal}{\emph{Philosophical Transactions of the Royal
  Society B: Biological Sciences}} \bibinfo{volume}{364},
  \bibinfo{number}{1521} (\bibinfo{year}{2009}), \bibinfo{pages}{1211--1221}.
\newblock


\bibitem[\protect\citeauthoryear{Huang and Rao}{Huang and Rao}{2011}]%
        {huang2011predictive}
\bibfield{author}{\bibinfo{person}{Yanping Huang} {and}
  \bibinfo{person}{Rajesh~PN Rao}.} \bibinfo{year}{2011}\natexlab{}.
\newblock \showarticletitle{Predictive coding}.
\newblock \bibinfo{journal}{\emph{Wiley Interdisciplinary Reviews: Cognitive
  Science}} \bibinfo{volume}{2}, \bibinfo{number}{5} (\bibinfo{year}{2011}),
  \bibinfo{pages}{580--593}.
\newblock


\bibitem[\protect\citeauthoryear{Keller and Mrsic-Flogel}{Keller and
  Mrsic-Flogel}{2018}]%
        {keller2018predictive}
\bibfield{author}{\bibinfo{person}{Georg~B Keller} {and}
  \bibinfo{person}{Thomas~D Mrsic-Flogel}.} \bibinfo{year}{2018}\natexlab{}.
\newblock \showarticletitle{Predictive processing: a canonical cortical
  computation}.
\newblock \bibinfo{journal}{\emph{Neuron}} \bibinfo{volume}{100},
  \bibinfo{number}{2} (\bibinfo{year}{2018}), \bibinfo{pages}{424--435}.
\newblock


\bibitem[\protect\citeauthoryear{Kok and de~Lange}{Kok and de~Lange}{2015}]%
        {kok2015predictive}
\bibfield{author}{\bibinfo{person}{Peter Kok} {and} \bibinfo{person}{Floris~P
  de Lange}.} \bibinfo{year}{2015}\natexlab{}.
\newblock \showarticletitle{Predictive coding in sensory cortex}.
\newblock In \bibinfo{booktitle}{\emph{An introduction to model-based cognitive
  neuroscience}}. \bibinfo{publisher}{Springer}, \bibinfo{pages}{221--244}.
\newblock


\bibitem[\protect\citeauthoryear{Kolen and Pollack}{Kolen and Pollack}{1994}]%
        {kolen1994backpropagation}
\bibfield{author}{\bibinfo{person}{John~F Kolen} {and}
  \bibinfo{person}{Jordan~B Pollack}.} \bibinfo{year}{1994}\natexlab{}.
\newblock \showarticletitle{Backpropagation without weight transport}. In
  \bibinfo{booktitle}{\emph{Proceedings of 1994 IEEE International Conference
  on Neural Networks (ICNN'94)}}, Vol.~\bibinfo{volume}{3}. IEEE,
  \bibinfo{pages}{1375--1380}.
\newblock


\bibitem[\protect\citeauthoryear{Kunin, Nayebi, Sagastuy-Brena, Ganguli, Bloom,
  and Yamins}{Kunin et~al\mbox{.}}{2020}]%
        {kunin2020two}
\bibfield{author}{\bibinfo{person}{Daniel Kunin}, \bibinfo{person}{Aran
  Nayebi}, \bibinfo{person}{Javier Sagastuy-Brena}, \bibinfo{person}{Surya
  Ganguli}, \bibinfo{person}{Jonathan Bloom}, {and} \bibinfo{person}{Daniel
  Yamins}.} \bibinfo{year}{2020}\natexlab{}.
\newblock \showarticletitle{Two routes to scalable credit assignment without
  weight symmetry}. In \bibinfo{booktitle}{\emph{International Conference on
  Machine Learning}}. PMLR, \bibinfo{pages}{5511--5521}.
\newblock


\bibitem[\protect\citeauthoryear{Lillicrap, Cownden, Tweed, and
  Akerman}{Lillicrap et~al\mbox{.}}{2016}]%
        {lillicrap2016random}
\bibfield{author}{\bibinfo{person}{Timothy~P Lillicrap},
  \bibinfo{person}{Daniel Cownden}, \bibinfo{person}{Douglas~B Tweed}, {and}
  \bibinfo{person}{Colin~J Akerman}.} \bibinfo{year}{2016}\natexlab{}.
\newblock \showarticletitle{Random synaptic feedback weights support error
  backpropagation for deep learning}.
\newblock \bibinfo{journal}{\emph{Nature communications}} \bibinfo{volume}{7},
  \bibinfo{number}{1} (\bibinfo{year}{2016}), \bibinfo{pages}{1--10}.
\newblock


\bibitem[\protect\citeauthoryear{Millidge, Tschantz, and Buckley}{Millidge
  et~al\mbox{.}}{2020}]%
        {millidge2020predictive}
\bibfield{author}{\bibinfo{person}{Beren Millidge}, \bibinfo{person}{Alexander
  Tschantz}, {and} \bibinfo{person}{Christopher~L Buckley}.}
  \bibinfo{year}{2020}\natexlab{}.
\newblock \showarticletitle{Predictive coding approximates backprop along
  arbitrary computation graphs}.
\newblock \bibinfo{journal}{\emph{arXiv preprint arXiv:2006.04182}}
  (\bibinfo{year}{2020}).
\newblock


\bibitem[\protect\citeauthoryear{Mostafa, Ramesh, and Cauwenberghs}{Mostafa
  et~al\mbox{.}}{2018}]%
        {mostafa2018deep}
\bibfield{author}{\bibinfo{person}{Hesham Mostafa}, \bibinfo{person}{Vishwajith
  Ramesh}, {and} \bibinfo{person}{Gert Cauwenberghs}.}
  \bibinfo{year}{2018}\natexlab{}.
\newblock \showarticletitle{Deep supervised learning using local errors}.
\newblock \bibinfo{journal}{\emph{Frontiers in neuroscience}}
  \bibinfo{volume}{12} (\bibinfo{year}{2018}), \bibinfo{pages}{608}.
\newblock


\bibitem[\protect\citeauthoryear{Neftci, Mostafa, and Zenke}{Neftci
  et~al\mbox{.}}{2019}]%
        {neftci2019surrogate}
\bibfield{author}{\bibinfo{person}{Emre~O Neftci}, \bibinfo{person}{Hesham
  Mostafa}, {and} \bibinfo{person}{Friedemann Zenke}.}
  \bibinfo{year}{2019}\natexlab{}.
\newblock \showarticletitle{Surrogate gradient learning in spiking neural
  networks: Bringing the power of gradient-based optimization to spiking neural
  networks}.
\newblock \bibinfo{journal}{\emph{IEEE Signal Processing Magazine}}
  \bibinfo{volume}{36}, \bibinfo{number}{6} (\bibinfo{year}{2019}),
  \bibinfo{pages}{51--63}.
\newblock


\bibitem[\protect\citeauthoryear{Qiao, Mostafa, Corradi, Osswald, Stefanini,
  Sumislawska, and Indiveri}{Qiao et~al\mbox{.}}{2015}]%
        {qiao2015reconfigurable}
\bibfield{author}{\bibinfo{person}{Ning Qiao}, \bibinfo{person}{Hesham
  Mostafa}, \bibinfo{person}{Federico Corradi}, \bibinfo{person}{Marc Osswald},
  \bibinfo{person}{Fabio Stefanini}, \bibinfo{person}{Dora Sumislawska}, {and}
  \bibinfo{person}{Giacomo Indiveri}.} \bibinfo{year}{2015}\natexlab{}.
\newblock \showarticletitle{A reconfigurable on-line learning spiking
  neuromorphic processor comprising 256 neurons and 128K synapses}.
\newblock \bibinfo{journal}{\emph{Frontiers in neuroscience}}
  \bibinfo{volume}{9} (\bibinfo{year}{2015}), \bibinfo{pages}{141}.
\newblock


\bibitem[\protect\citeauthoryear{Rao and Ballard}{Rao and Ballard}{1999a}]%
        {Rao_Ballard99_predcodi}
\bibfield{author}{\bibinfo{person}{Rajesh~PN Rao} {and} \bibinfo{person}{Dana~H
  Ballard}.} \bibinfo{year}{1999}\natexlab{a}.
\newblock \showarticletitle{Predictive coding in the visual cortex: a
  functional interpretation of some extra-classical receptive-field effects}.
\newblock \bibinfo{journal}{\emph{Nature neuroscience}} \bibinfo{volume}{2},
  \bibinfo{number}{1} (\bibinfo{year}{1999}), \bibinfo{pages}{79--87}.
\newblock


\bibitem[\protect\citeauthoryear{Rao and Ballard}{Rao and Ballard}{1999b}]%
        {rao1999predictive}
\bibfield{author}{\bibinfo{person}{Rajesh~PN Rao} {and} \bibinfo{person}{Dana~H
  Ballard}.} \bibinfo{year}{1999}\natexlab{b}.
\newblock \showarticletitle{Predictive coding in the visual cortex: a
  functional interpretation of some extra-classical receptive-field effects}.
\newblock \bibinfo{journal}{\emph{Nature neuroscience}} \bibinfo{volume}{2},
  \bibinfo{number}{1} (\bibinfo{year}{1999}), \bibinfo{pages}{79--87}.
\newblock


\bibitem[\protect\citeauthoryear{Song, Lukasiewicz, Xu, and Bogacz}{Song
  et~al\mbox{.}}{2020}]%
        {song2020can}
\bibfield{author}{\bibinfo{person}{Yuhang Song}, \bibinfo{person}{Thomas
  Lukasiewicz}, \bibinfo{person}{Zhenghua Xu}, {and} \bibinfo{person}{Rafal
  Bogacz}.} \bibinfo{year}{2020}\natexlab{}.
\newblock \showarticletitle{Can the brain do backpropagation? exact
  implementation of backpropagation in predictive coding networks}.
\newblock \bibinfo{journal}{\emph{Advances in neural information processing
  systems}}  \bibinfo{volume}{33} (\bibinfo{year}{2020}),
  \bibinfo{pages}{22566}.
\newblock


\bibitem[\protect\citeauthoryear{Spratling}{Spratling}{2019}]%
        {spratling2019fitting}
\bibfield{author}{\bibinfo{person}{MW Spratling}.}
  \bibinfo{year}{2019}\natexlab{}.
\newblock \showarticletitle{Fitting predictive coding to the neurophysiological
  data}.
\newblock \bibinfo{journal}{\emph{Brain research}}  \bibinfo{volume}{1720}
  (\bibinfo{year}{2019}), \bibinfo{pages}{146313}.
\newblock


\bibitem[\protect\citeauthoryear{Spratling}{Spratling}{2008}]%
        {spratling2008predictive}
\bibfield{author}{\bibinfo{person}{Michael~W Spratling}.}
  \bibinfo{year}{2008}\natexlab{}.
\newblock \showarticletitle{Predictive coding as a model of biased competition
  in visual attention}.
\newblock \bibinfo{journal}{\emph{Vision research}} \bibinfo{volume}{48},
  \bibinfo{number}{12} (\bibinfo{year}{2008}), \bibinfo{pages}{1391--1408}.
\newblock


\bibitem[\protect\citeauthoryear{Spratling, De~Meyer, and Kompass}{Spratling
  et~al\mbox{.}}{2009}]%
        {spratling2009unsupervised}
\bibfield{author}{\bibinfo{person}{Michael~W Spratling}, \bibinfo{person}{Kris
  De~Meyer}, {and} \bibinfo{person}{R Kompass}.}
  \bibinfo{year}{2009}\natexlab{}.
\newblock \showarticletitle{Unsupervised learning of overlapping image
  components using divisive input modulation}.
\newblock \bibinfo{journal}{\emph{Computational intelligence and neuroscience}}
   \bibinfo{volume}{2009} (\bibinfo{year}{2009}).
\newblock


\bibitem[\protect\citeauthoryear{Wacongne, Changeux, and Dehaene}{Wacongne
  et~al\mbox{.}}{2012}]%
        {wacongne2012neuronal}
\bibfield{author}{\bibinfo{person}{Catherine Wacongne},
  \bibinfo{person}{Jean-Pierre Changeux}, {and} \bibinfo{person}{Stanislas
  Dehaene}.} \bibinfo{year}{2012}\natexlab{}.
\newblock \showarticletitle{A neuronal model of predictive coding accounting
  for the mismatch negativity}.
\newblock \bibinfo{journal}{\emph{Journal of Neuroscience}}
  \bibinfo{volume}{32}, \bibinfo{number}{11} (\bibinfo{year}{2012}),
  \bibinfo{pages}{3665--3678}.
\newblock


\bibitem[\protect\citeauthoryear{Walsh, McGovern, Clark, and O'Connell}{Walsh
  et~al\mbox{.}}{2020}]%
        {walsh2020evaluating}
\bibfield{author}{\bibinfo{person}{Kevin~S Walsh}, \bibinfo{person}{David~P
  McGovern}, \bibinfo{person}{Andy Clark}, {and} \bibinfo{person}{Redmond~G
  O'Connell}.} \bibinfo{year}{2020}\natexlab{}.
\newblock \showarticletitle{Evaluating the neurophysiological evidence for
  predictive processing as a model of perception}.
\newblock \bibinfo{journal}{\emph{Annals of the new York Academy of Sciences}}
  \bibinfo{volume}{1464}, \bibinfo{number}{1} (\bibinfo{year}{2020}),
  \bibinfo{pages}{242}.
\newblock


\bibitem[\protect\citeauthoryear{Whittington and Bogacz}{Whittington and
  Bogacz}{2017}]%
        {whittington2017approximation}
\bibfield{author}{\bibinfo{person}{James~CR Whittington} {and}
  \bibinfo{person}{Rafal Bogacz}.} \bibinfo{year}{2017}\natexlab{}.
\newblock \showarticletitle{An approximation of the error backpropagation
  algorithm in a predictive coding network with local hebbian synaptic
  plasticity}.
\newblock \bibinfo{journal}{\emph{Neural computation}} \bibinfo{volume}{29},
  \bibinfo{number}{5} (\bibinfo{year}{2017}), \bibinfo{pages}{1229--1262}.
\newblock


\bibitem[\protect\citeauthoryear{Xiao, Chen, Liao, and Poggio}{Xiao
  et~al\mbox{.}}{2018}]%
        {xiao2018biologically}
\bibfield{author}{\bibinfo{person}{Will Xiao}, \bibinfo{person}{Honglin Chen},
  \bibinfo{person}{Qianli Liao}, {and} \bibinfo{person}{Tomaso Poggio}.}
  \bibinfo{year}{2018}\natexlab{}.
\newblock \showarticletitle{Biologically-plausible learning algorithms can
  scale to large datasets}.
\newblock \bibinfo{journal}{\emph{arXiv preprint arXiv:1811.03567}}
  (\bibinfo{year}{2018}).
\newblock


\end{thebibliography}

\appendix

\section{Research Methods}

\subsection{Division Encoding Cost Function}

The division encoding error $e^{**}_l$ is equal to $\sqrt{\frac{a_l}{f(p_l)}}$. Under this encoding scheme, $a_l = p_l$ when $e^{**}_l = 1$. The goal then is to update weights and activities to reduce the difference between $e^{**}_l$ and 1. To do so, we use the following cost function:
\begin{equation}
C = \sum_l^L C_l,\text{ where } C_l = \frac12 \log(e^{**}_l)^2.
\end{equation}
When $e^{**}_l = 1$ the log of $e^{**}_l$ will equal 0 so the cost at level $l$ will equal 0. Deviations from 1, will lead the cost to increase. We use the log here, instead of simply subtracting $e^{**}_l$ from 1 because it simplifies the computations of the gradients.

\subsubsection{Weight Updates for Division Encoding}
Here we derive the gradients of C at level $l$ with respect to the forward weights at level $l$:
\begin{displaymath}
\begin{split}
C_l & = \frac12 (\log(e^{**}_l))^2,\\
& = \frac12 (\frac12(\log(a_l) - \log(f(p_l))))^2.
\end{split}
\end{displaymath}
Lets call the term inside the parentheses $u_l$, such that $C_l = \frac12 u_l^2$. Now we need to compute $\frac{\partial C_l}{\partial W_{l-1}}$, which can be decomposed using the chain rule as follows.

\begin{displaymath}
\frac{\partial C_l}{\partial W_l} = \frac{\partial C_l}{\partial u_l} \frac{\partial u_l}{\partial f(p_l)}\, \frac{\partial f(p_l)}{\partial p_l} \frac{\partial p_l}{\partial W_{l-1}}. 
\end{displaymath}
Each term can then be computed individually.
\begin{displaymath}
\begin{split}
\frac{\partial C_l}{\partial u_l} & = u_l = \frac12(\log(a_l) - \log(f(p_l))),\\
\frac{\partial u_l}{\partial f(p_1)} & =  -1 / f(p_l),\\
\frac{\partial f(p_l)}{\partial p_l} & = f'(p_l),\\
\frac{\partial p_l}{\partial W_{l-1}} & = a_{l-1}^T.
\end{split}
\end{displaymath}
Now if we combine terms we get the following weight update
\begin{equation}
\begin{split}
\frac{\partial C_l}{\partial W_l} & = \frac12(\log(a_l) - \log(p_l)) (\frac{-1}{f(p_l)}) f'(p_l) a_{l-1}^T, \\
& = -\log(\sqrt{\frac{a_l}{f(p_l)}}) \frac{f'(p_l)}{f(p_l)} a_{l-1}^T,\\
& = -\log(e_l^{**}) \frac{f'(p_l)}{f(p_l)} a_{l-1}^T.
\end{split}
\end{equation}
We can see this update is the outer product of pre-synaptic and post-synaptic information. In particular, the pre-synaptic activity is multiplied by the error neuron activities, which are first passed through a non-linearity (logarithm) and multiplied by information about the predictions at the same level.

\subsubsection{Activity Updates for Division Encoding}

Here we derive the gradients of the cost at level $l$ w.r.t. the activities. Like original equation, the gradients w.r.t. $a_l$ are derived from the cost at the same level and the next level (i.e. level $l$ and $l+1$). The gradient of $C_{l+1}$ w.r.t $a_l$ can be decomposed using the chain rule as follows.
\begin{displaymath}
\frac{\partial C_{l+1}}{\partial a_{l}} = \frac{\partial C_{l+1}}{\partial u_{l+1}} \frac{\partial u_{l+1}}{\partial f(p_{l+1})} \frac{\partial f(p_{l+1})}{\partial p_{l+1}} \frac{\partial p_{l+1}}{\partial a_l}.
\end{displaymath}
We already computed all of these terms in the last section except for $\frac{\partial p_l}{\partial a_{l-1}} $. The gradient $\frac{\partial p_l}{\partial a_{l-1}}$ is just equal to $W^T_l$. This gives us the bottom-up error (the first term in equation 4) for the activity update $a_l$:
\begin{displaymath}
\frac{\partial C_{l+1}}{\partial a_{l}} = W_{l}^T \log(e_l^{**}) \frac{f'(p_l)}{f(p_l)}. 
\end{displaymath}
Now we need to compute $\frac{\partial C_{l}}{\partial a_{l}}$. This term can be decomposed using the chain rule as follows.
\begin{displaymath}
\frac{\partial C_{l}}{\partial a_{l}} = \frac{\partial C_{l}}{\partial u_{l}} \frac{\partial u_{l}}{\partial a_l}.
\end{displaymath}
We saw in the last section that $\frac{\partial C_l}{\partial u_l}$ is equal to $\log(e_l^{**})$. $\frac{\partial u_{l}}{\partial a_l}$ is $\frac{1}{a_l}$. With these terms we can now compute the top-down error for the activity update (second term in equation 4):
\begin{displaymath}
\frac{\partial C_{l}}{\partial a_{l}} = \log(e_l^{**}) \frac{1}{a_l}
\end{displaymath}
Combining the top-down and bottom up error now gets us the activity updates under the division encoding schema:
\begin{equation}
\Delta a_l = W_{l}^T \log(e_l^{**}) \frac{f'(p_l)}{f(p_l)} - \log(e_l^{**}) \frac{1}{a_l}
\end{equation}
\end{document}